\documentclass[11pt]{article}
\usepackage{coling2020}
\usepackage{times}
\usepackage{url}
\usepackage{latexsym}

\usepackage{microtype}
\usepackage{amsthm}
\usepackage{algorithm}
\usepackage{helvet}
\usepackage{courier}
\usepackage{textcomp}
\usepackage{enumerate}
\usepackage{tabularx}
\usepackage{multirow}
\usepackage{array}
\usepackage{graphics}
\usepackage{graphicx}
\usepackage{subfig}
\usepackage{color}
\usepackage{amsmath}
\usepackage{makecell}
\usepackage{indentfirst}
\usepackage{multirow}
\usepackage{amssymb}
\usepackage{algpseudocode}
\usepackage{comment}
\usepackage{mathrsfs}
\usepackage{extarrows}
\usepackage{booktabs}

\colingfinalcopy 

\title{An Enhanced Knowledge Injection Model for Commonsense Generation}

\author{Zhihao Fan$^{1}$\footnotemark[1], Yeyun Gong$^{2}$, Zhongyu Wei$^{1,5}$\footnotemark[2], Siyuan Wang$^{1}$, \textbf{Yameng Huang}$^{3}$, \\
\textbf{Jian Jiao}$^{3}$\textbf{, Xuanjing Huang}$^{4}$\textbf{, Nan Duan}$^{2}$\textbf{, Ruofei Zhang}$^{3}$\\
$^{1}$School of Data Science, Fudan University, China\\
$^{2}$Microsoft Research Asia, $^{3}$Microsoft \\
$^{4}$School of Computer Science, Fudan University, China\\
$^{5}$Research Institute of Intelligent and Complex Systems, Fudan University, China\\
\{fanzh18,zywei,wangsy18,xjhuang\}@fudan.edu.cn, \\
\{yegong,yameng.huang,Jian.Jiao,nanduan,bzhang\}@microsoft.com
}

\begin{document}
\maketitle
\begin{abstract}
    Commonsense generation aims at generating plausible everyday scenario description based on a set of provided concepts. Digging the relationship of concepts from scratch is non-trivial, therefore, we retrieve prototypes from external knowledge to assist the understanding of the scenario for better description generation. We integrate two additional modules, namely position indicator and scaling module, into the pretrained encoder-decoder model for prototype modeling to enhance the knowledge injection procedure. We conduct experiment on CommonGen benchmark, and experimental results show that our method significantly improves the performance on all the metrics.
\end{abstract}


\section{Introduction}
\label{intro}
\blfootnote{
    %
    %
    %
    \hspace{-0.65cm}  
    This work is licensed under a Creative Commons 
    Attribution 4.0 International License.
    License details:
    \url{http://creativecommons.org/licenses/by/4.0/}.
}

Recently, commonsense reasoning tasks~\cite{zellers2018swag,talmor2018commonsenseqa,lin2019comgen} have been proposed to investigate the ability of machines to make acceptable and logical inferences about ordinary scenes in our daily life. Both SWAG~\cite{zellers2018swag} and CommonsenseQA~\cite{talmor2018commonsenseqa} present a piece of text (an event description or a question) together with several choices (subsequent events or answers), and the system is asked to choose the correct option based on the context. Different from these two discriminative tasks, CommonGen~\cite{lin2019comgen} moves to a generation setting. It requires the system to construct a logical sentence based on several concepts related to a specific scenario.



The task of text generation from given concepts are challenging in two ways. First, the sentence needs to be grammatically sound with the constraints of including given concepts. Second, the sentence needs to be correct in terms of our common knowledge. Existing approaches apply pretrained encoder-decoder models~\cite{lewis2019bart,bao2020unilmv2} for description construction and concepts are modeled as constraints to guide the generation process. Sentences generated by these models are fluent, however, the output might violates the commonsense. An example is shown in Table~\ref{comparison}. The model \emph{BART} generates sentence with ``guitar sits" which is incorrect. This demonstrates that the language model itself is not able to determine the rational relationship between concepts. 


\begin{table}[!th]
\begin{center}
\begin{tabular}{l|l|l}
\midrule[1.0pt]
\emph{Concepts} &front, guitar, microphone, sit &ear, feel, pain, pierce  \\
\midrule[1.0pt]
\multirow{2}{*}{\emph{BART}}  &\underline{guitar sits} in front of a microphone  &I can feel the pain in my ears and \underline{feel} \\
&in the front. &\underline{the pierce} in my neck from the piercing. \\
\midrule[0.5pt]
\multirow{2}{*}{\emph{Prototype}} &A singer performed the song standing in  &He expresses severe pain as he tries \\
&front of the audiences while playing guitar. &to pierce his hand.\\
\midrule[0.5pt]
\emph{BART+}&A singer sitting in front of the audiences &He expresses severe pain as he \\
\emph{Prototype}&while playing guitar. &pierces his ear.\\
\midrule[1.0pt]
\end{tabular}
\end{center}
\caption{Example of \emph{BART}, \emph{Prototype} and \emph{BART+Prototype}.}
\label{comparison}
\end{table}

In order to enrich the source information and bridge the semantic gap between source and target, we argue that external knowledge related to the scene of given concepts are needed to determine the relationships between concepts. Motivated by the retrieve-and-generation framework~\cite{song2016two,hashimoto2018retrieve} for text generation, we retrieve prototypes for concepts from external corpora as scene knowledge and construct sentences by editing prototypes. The prototype would introduce scenario knowledge to make up the shortcoming of the language model in finding out reasonable concept combination. Furthermore, prototypes would provide the missing key concepts besides the provided concept set, such as ``singer'' of the first example in Table~\ref{comparison}, to help complete a natural and coherent scenario.

In order to better utilize the prototypes, we propose two additional modules on top of the pretrained encoder-decoder model with the guidance of given concepts. First, considering tokens in the prototype make various contributions in the sentence generation, a \emph{scaling module} is introduced to assign weights to tokens in the prototype. Second, tokens closer to the concept words in prototypes might be more important for scene description generation, therefore, a \emph{position indicator} is proposed to mark the relative position of different tokens in the prototypes. The main contributions of this work are three folds. 1) We propose a retrieve-and-edit framework, \textbf{E}nhanced \textbf{K}nowledge \textbf{I}njection \textbf{BART}, for the task of commonsense generation. 2) We combine the two modules, scaling module and prototype position indicator, to better utilize the scenario knowledge of prototype.  3) we conduct experiments on CommonGen benchmark, and results show that our method achieves significantly improvement by using both in-domain and out-of-domain plain text datasets as external knowledge source.

\section{Model}

\begin{figure}[t]
    \centering
    \includegraphics[width = 4.5in]{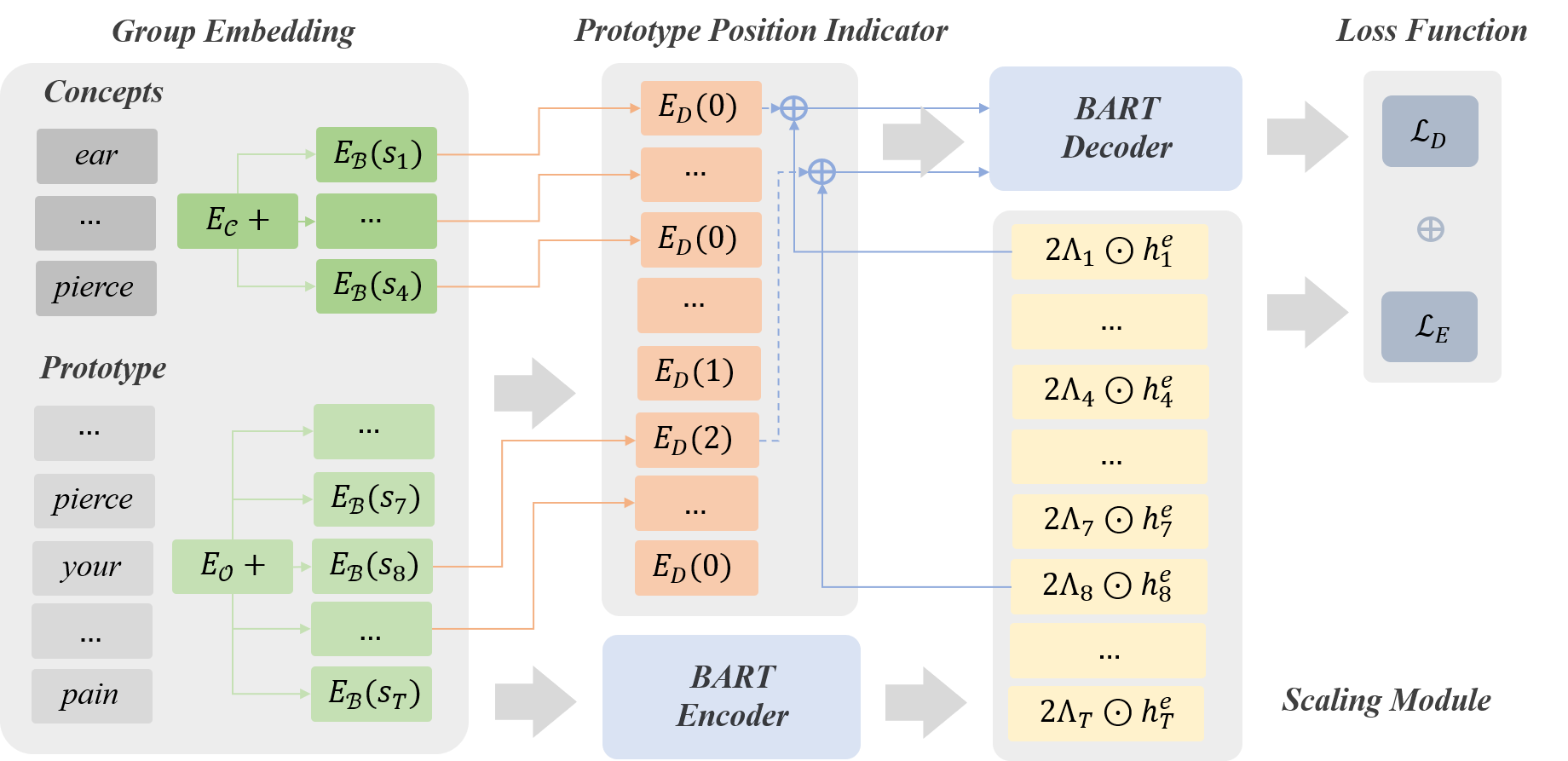}
	\caption{The framework of our proposed \emph{EKI-BART}. $E_{B}$ , $\mathit{E}_{\mathcal{C}}$, $\mathit{E}_{\mathcal{O}}$ and $\mathit{E}_{D}$ are the embedding function of \emph{BART} model, concept $\mathcal{C}$, prototype $\mathcal{O}$ and distance of prototype position indicator. $s_{v}$ and $h^{e}_{v}$ are the $v$th token of input and the corresponding \emph{BART} encoder output. $\mathcal{L}_{E}$ and $\mathcal{L}_{D}$ are classification loss and the loss-likelihood loss, respectively. Refers to Table~\ref{comparison} for the example in the framework.}
	\label{framework}
\end{figure}

In this section, we introduce our retrieve-and-generation framework based \emph{EKI-BART} as $G_{\theta}$ with parameter $\theta$ that retrieves prototype $\mathcal{O}=(o_{1},o_{2},\cdots,o_{n_{o}})$ from external text knowledge corpus and extracts the prototype knowledge under the guidance of concepts $\mathcal{C}=(c_{1},\cdots,c_{n_{c}})$ to improve the commonsense generation of target $\mathcal{T}=(t_{1},\cdots,t_{n_{t}})$. 
The overall framework of our proposed model is shown in Figure~\ref{framework}.

\subsection{Pretrained Encoder-Decoder}
Pretrained encoder-decoder model, \emph{BART}~\cite{lewis2019bart}, commonly follows the transformer architecture. Several encoder-layers stack as encoder and each of them is composed of a self-attention network and a feed-forward network. The input sequence would be encoded into a hidden state sequence $\mathcal{H}^{e}=(h^{e}_{1},\cdots,h^{e}_{n_{h}})$. Decoder is also stacked by a few decoder-layers and the key difference between encoder-layer and decoder-layer is that there exists an encoder-decoder-attention in the middle of self-attention and feed-forward network. In each encoder-decoder-attention module, the decoder representation $h^{d}_{u}$ would attend to $\mathcal{H}^{e}$ following Equation~\ref{raw-decoder-encoder-attention}.
\begin{align}
    &s_{x}(h^{d}_{u}, h^{e}_{v})=(W_{x,q}h^{d}_{u})^{T}(W_{x,k}h^{e}_{v})\big/\sqrt{d_k} \nonumber\\
    &a_{x}=softmax\big(s_{x}(h^{d}_{u},h^{e}_{1}),\cdots,s_{x}(h^{d}_{u},h^{e}_{n_{h}})\big) \nonumber \\
    &\hat{h}^{d}_{u}=W_{o}\big[W_{1,v}\mathcal{H}^{e}a_{1},\cdots,W_{X,v}\mathcal{H}^{e}a_{X}\big] \nonumber \\
    &h^{d}_{u}=LN\big(h^{d}_{u}+\hat{h}^{d}_{u}\big)
    ~\label{raw-decoder-encoder-attention}
\end{align}
where $x$ denotes the $x$th attention head, where $\{W_{x,q},W_{x,k},W_{x,v}\}\in \mathbb{R}^{d_{k}\times d}$ are trainable parameters for query, key and value, $x$ denotes the attention head, $d$ is the hidden size, $d_k$ is the attention head dimension, and $LN$ is the layernorm function. Generally, there is a normalization operation before we get the encoder output, in other words, the correlation between $h^{e}_{v}$ and $h^{d}_{u}$ mainly depends on the direction of $h^{e}_{v}$ and $h^{d}_{u}$.


\subsection{Model Input}
Following the input setting of \emph{BART}, we concatenate the provided concepts $\mathcal{C}$ and the retrieved prototype $\mathcal{O}$ as a whole input $\mathcal{S}$ to feed into the pretrained model.
\begin{gather}
    \mathcal{S}=\big[\mathcal{C},\mathcal{O}\big]=\big[c_{1},\cdots,c_{n_{c}},o_{1},\cdots,o_{n_{o}}\big]
\end{gather}
where $\big[\cdot,\cdot\big]$ is the concatenation operation.

In our retrieve-and-generation framework, we need to modify the prototype $\mathcal{O}$ to meet the requirement of $\mathcal{C}$. To distinguish each token from $\mathcal{O}$ or $\mathcal{C}$, we add the group embedding on top of original \emph{BART} embedding function as Equation~\ref{EMB} shows.
\begin{align}
    \mathit{E}(c_{j})=\mathit{E}_{B}(c_{j})+\mathit{E}_{\mathcal{C}},\
    \mathit{E}(o_{k})=\mathit{E}_{B}(o_{k})+\mathit{E}_{\mathcal{O}} \label{EMB}
\end{align}
where $\mathit{E}_{B}$ stands for the original embedding function in \emph{BART} including token embedding and position embedding, $\mathit{E}_{\mathcal{C}}$ and $\mathit{E}_{\mathcal{O}}$ are two group embedding for concepts $\mathcal{C}$ and prototype ${\mathcal{O}}$, and $\mathit{E}$ is the final embedding function. 

\subsection{Generation}
The prototype $\mathcal{O}$ not only introduces scenario bias and effective additional concepts, but also brings noises into generation. In order to inject the retrieved knowledge into generation more effectively, we argue to extract the scenario knowledge of prototype in a more fine-grained manner. From Equation~\ref{raw-decoder-encoder-attention}, we can see that each token in $\mathcal{S}$ gets involved in encoder-decoder-attention with the encoder output $h^{e}_{v}$, thus we propose two mechanisms, namely, scaling module and prototype position indicator, to improve the generation. 


\subsubsection{Encoder with Scaling Module}
We observe that noises and concept tokens both appear in the retrieved prototype, and these noises would dominate the generation.
The simplest solution is to utilize a hard mask, in other words, only keep those concept tokens in prototype and mask others, but the decoder would be no longer aware of the complete prototype scenario, and these effective additional concepts would be also unavailable. Instead of hard masking, we propose scaling module to assign scaling factor for input tokens which can be applied in encoder-decoder-attention, then the model is capable of receiving less noises and learn more from effective tokens.

We investigate the dot product based attention mechanism shown in Equation~\ref{raw-decoder-encoder-attention}. Function $\mathit{F}$ with a scaling factor $\lambda$ on top of the normalized encoder output states $\mathcal{H}$ is defined in Equation~\ref{scale1},
\begin{align}
    F(\lambda)&=S(h^{d}_{u}, \lambda h^{e}_{v})=\lambda \Big(\big(W_{q}h^{d}_{u}\big)^{T}\big(W_{k}h^{e}_{v}\big)\big/\sqrt{d_k}\Big)=\lambda S(h^{d}_{u}, h^{e}_{v})=\lambda F(1) ~\label{scale1}
\end{align}
From Equation~\ref{scale1}, we can see that when $\big(W_{q}h^{d}_{u}\big)^{T}\big(W_{k}h^{e}_{v}\big)$ is a large positive value or $h^{e}_{v}$ takes important attention weights in $h^{d}_{u}$, then $F(\lambda)$ is a monotonically decreasing function. This inspires us to refine the representation of $h^{e}_{v}$ through $\lambda$. Viewing $\lambda$ as an importance factor, we are able to weaken/strength $h^{e}_{v}$ in encoder-decoder-attention through decreasing/increasing $\lambda$.

With the awareness of the phenomenon in Equation~\ref{scale1},  we devise a scaling module on the basis of Equation~\ref{raw-decoder-encoder-attention}.  
In practice, we attach a scaling module to the encoder, which can increase the norm if $h^{e}_{v}$ is likely to contribute to the generation and decrease when the $h^{e}_{v}$ has a conflict with concepts. Each channel of $h^{e}_{v}$ would be taken into account separately. This is accomplished with the following scaling module. The module is composed of 
\begin{gather}
    \Lambda=\mathop{Sigmoid}\Big(W_{2}ReLU\big(W_{1}h^{e}_{v}+b_{1}\big)+b_{2}\Big) \nonumber \\
    h^{e}_{v}=h^{e}_{v}\odot\big(2\times\Lambda\big) \label{scale-output}
\end{gather}
where $W_{1}\in\mathbb{R}^{d_{s}\times d},W_{2}\in\mathbb{R}^{d\times d_{s}},b_{1}\in\mathbb{R}^{d_{s}},b_{2}\in\mathbb{R}^{d}$ are trainable parameters in the scaling module.

Consider that the parameters of pretrained encoder-decoder model have been optimized during pretraining, simply adding the parameter $\Lambda$ may destroy the distribution of encoder output $\mathcal{H}$ and leads to training failure. We try to initialize these parameters in scaling module with $N(0,var)$, where $var$ is a small value, then the output with sigmoid activation would gather around 0.5, and $2\times$ would make them fall near 1.0. Thus in the beginning of training, the participation of scaling module would not lead to a mess.

In our knowledge, prototype tokens that co-occur in $\mathcal{T}$ should be more important than others for the generation of $\mathcal{T}$. We hope these prior knowledge could help the model to better discriminate the importance of these prototype tokens, thus we introduce an encoder classification task that requires the scaling module to determine which tokens would appear in the generated sentence. 
\begin{gather}
    \mathcal{L}_{E}=-\sum_{s_{v}\in\mathcal{S}}\Big(\mathcal{I}_{\{s_{v}\in\mathcal{T}\}}\log\mathop{Mean}(\Lambda_{v})+\mathcal{I}_{\{s_{v}\notin\mathcal{T}\}}\log\big(1-\mathop{Mean}(\Lambda_{v})\big)\Big)
\end{gather}
where $Mean$ is to get the mean value and $\mathcal{I}$ is indicator function, $\mathcal{I}_{\{s_{v}\in\mathcal{T}\}}=0$ if $s_{v}\in\mathcal{T}$ otherwise 1, so is $\mathcal{I}_{\{s_{v}\notin\mathcal{T}\}}$.

\subsubsection{Decoder with Prototype Position Indicator}
These surrounding tokens of concept tokens in prototype $\mathcal{O}$ tend to describe how these concepts interact with the complete scenario. We argue that informing the decoder of these relative positions would help decoder better learn effective scenario bias of the prototype $\mathcal{O}$.



Before the computation of encoder-decoder-attention, we devise a position indicator function to assign positions to those tokens in prototype. First, we assign virtual positions to tokens in prototype $\mathcal{O}$ in sequence, from 1 to $n_{o}$. Second, we pick up the positions of those concept tokens in prototype as multiple position centers. Third, for each token $o_{v}\in\mathcal{O}$, we compute the smallest distance from $o_{v}$ to those concept tokens. The process is shown in Equation~\ref{dist_compute}.
\begin{gather}
    \mathit{D}(s_{v})=min\big\{|v-p|,s_{p}=c,s_{p}\in\mathcal{O},c\in \mathcal{C}\big\}\label{dist_compute}
\end{gather}

Our inputs tokens are composed of prototype ones and concept ones. Considering the particularity of concept words $\mathcal{C}$, we assign them with a default position value 0 and adjust the position indicator function of prototype tokens through adding one, the process is shown in Equation~\ref{all_dist_compute}.
\begin{gather}
\mathit{D}(s_{v})=\left\{\begin{array}{ll}
     \mathit{D}(s_{v})+1 & s_{v}\in \mathcal{O} \\
     0 & s_{v} \in \mathcal{C} 
\end{array} \right.\label{all_dist_compute}
\end{gather}

On the basis of the prototype position indicator function in Equation~\ref{all_dist_compute}, we add the information of relative position from each token itself to the closest concept token in prototype into encoder-decoder-attention through Equation~\ref{new-encoder-decoder-attention}.
\begin{gather}
    \mathit{ED}(h^{e}_{v})=\mathit{E}_{\mathit{D}}\big(\mathit{D}(s_{v})\big)\nonumber \\
    S(h^{d}_{u}, h^{e}_{v})=\big(W_{q}h^{d}_{u}\big)^{T}\big(W_{k}h^{e}_{v}+\mathit{ED}(h^{e}_{v})\big)\big/\sqrt{d_{k}}
    ~\label{new-encoder-decoder-attention}
\end{gather}
where $\mathit{E}_{\mathit{D}}$ is the embedding for those distance values in $\mathit{D}$. 
These prototype tokens that more close to the concept tokens are expected to receive more attention than other tokens.

\subsection{Training}
The objective of our model is to maximize the log-likelihood for $\mathcal{T}$ given $\mathcal{O}$ and $\mathcal{C}$.
\begin{gather}
    \mathcal{L}_{D}=-\mathop{log}\sum_{k}P(t_{k}|\mathcal{O},\mathcal{C},t_{<k})
\end{gather}
where $t_k$ in the $k$th token in $\mathcal{T}$ and $t_{<k}$ are the first $(k-1)$ tokens in $\mathcal{T}$.

During training, we try to minimize the sum of encoder classification loss and the decoder log-likelihood loss. The $\lambda$ is utilized to keep the balance between $\mathcal{L}_{D}$ and $\mathcal{L}_{E}$ such that encourages the model $G_{\theta}$ to achieve better performance in generation.
\begin{gather}
    \mathcal{L}=\mathcal{L}_{D}+\lambda\mathcal{L}_{E}\label{overall_loss}
\end{gather}

During prediction, we decode with beam search, and keep the sequence with highest predicted probability among those in the last beam.

\section{Experiment}
In this section, we conduct experiments on CommonGen benchmark to evaluate the effectiveness of our proposed approach. To dig into our approach, we perform ablation studies to explore the different effects of scaling module and prototype position indicator. 

\subsection{Prototype Collection}
\paragraph{In-Domain Corpus $D_{in}$}CommonGen is to describe a common scenario in our daily life. Datasets of image captioning or video captioning would contain more knowledge about spatial relations, object properties, physical rules, temporal event knowledge and social conventions that contribute to build the target scene contains these provided concepts. We utilize VaTeX~\cite{wang2019vatex}, SNLI~\cite{snliemnlp2015}, Activity~\cite{krishna2017dense} and the training set of CommonGen as the external plain text knowledge datasets and retrieve prototype according to the concepts appear in the sentence.

\paragraph{Out-of-Domain Corpus  $D_{out}$}In-domain corpus $D_{in}$ may only suitable for these description sentences for daily scenario and has difficulty in generalizing to other domains, thus we also employ Wikipedia as our external knowledge dataset to retrieve prototypes to test the generalization of our model.

\begin{table}[!th]
\begin{center}
\begin{tabular}{l|ccccc}
\midrule[1.0pt]
 &1 &2 &3 &4 &5\\
\midrule[1.0pt]
$\mathcal{D}_{in}$ &2,179 &17,664 &16,356 &2,538 &332 \\
$\mathcal{D}_{out}$ &3,009 &21,441 &12,278 &2,069 &272 \\
\midrule[1.0pt]
\end{tabular}
\end{center}
\caption{The number of retrieved prototypes whose concepts co-occur in ground truth sentence across different external knowledge datasets $\mathcal{D}_{in}$ and $\mathcal{D}_{out}$.}
\label{concept_number}
\end{table}

The number of retrieved concepts in prototypes that co-occur in ground truth sentence across different external knowledge datasets $\mathcal{D}_{in}$ and $\mathcal{D}_{out}$ is shown in Table~\ref{concept_number}. It is easy to conclude that we are able to retrieve more relevant prototypes from in-domain dataset $\mathcal{D}_{in}$ compare to out-of-domain dataset $\mathcal{D}_{out}$.

\begin{table}
\begin{center}
\begin{tabular}{l|cc|cc|c|c|c}
\midrule[1.0pt]
Model &\multicolumn{2}{c}{ROUGE-2/L} &\multicolumn{2}{c}{BLEU-3/4} &METEOR &CIDEr &SPICE \\
\midrule[1.0pt]
\emph{bRNN-CopyNet}
&2.90&19.25&5.50&2.00&12.70&3.99&10.60\\
\emph{Trans-CopyNet}
&2.28&14.04&4.30&2.00&9.10&2.31&7.50\\
\emph{MeanPooling-CopyNet}
&3.30&19.35&6.60&2.40&13.50&4.34&13.00\\
\emph{LevenTrans}
&5.74&21.24&8.80&4.00&13.30&3.72&14.00\\
\midrule[1.0pt]
\emph{GPT-2}
&16.47&38.01&28.70&19.40&24.40&11.06&24.50\\
\emph{BERT-Gen}
&19.78&40.93&33.20&23.10&28.50&13.31&28.30\\
\emph{UniLM}
&21.57 &41.96 &38.30 &27.50 &29.40 &14.92 &29.90\\
\emph{UniLM-v2}
&21.02 &42.41 &34.80 &24.30 &29.80 &14.61 &30.00\\
\emph{T5}
&21.71 &41.79 &38.10 &27.20 &30.00 &14.58 &30.60\\
\emph{BART}
&22.38 &41.44 &35.10 &24.90 &30.50 &13.32 &30.10 \\
\midrule[1.0pt]
\emph{Retrieve}$_{D_{out}}$ &7.84 &26.25 &12.70 &7.50 &18.40 &4.95 &15.00\\
\emph{BART}$_{D_{out}}$ &22.87 &43.77 &41.20 &30.30 &31.50 &15.82 &31.80  \\
\emph{EKI-BART}$_{D_{out}}$ &24.36 &45.42 &42.90 &32.10 &32.00 &16.80 &32.50  \\
\midrule[0.5pt]
\emph{Retrieve}$_{D_{in}}$ &18.49 &40.73 &35.00 &26.40 &29.90 &12.91 &27.90\\
\emph{BART}$_{D_{in}}$ &23.15 &44.71 &42.20 &32.40 &32.30 &16.43 &32.70 \\
\emph{EKI-BART}$_{D_{in}}$ &\textbf{25.43} &\textbf{46.53} &\textbf{46.00} &\textbf{36.10} &\textbf{33.80} &\textbf{17.80} &\textbf{33.40} \\
\midrule[1.0pt]
\end{tabular}
\end{center}
\caption{Overall performance of different models for CommonGen. Numbers in \textbf{bold} denotes the best performance in each column.}  
\label{OverallPerformance}
\end{table}

\subsection{Experimental Setup}
CommonGen~\cite{lin2019comgen} dataset contains 27,069, 993 and 1,497 concept-sets in training, validation and test set, the sentences are 39,069, 4,018 and 6,042 respectively. The proportion of novel concept-sets in validation and test datasets are $95.53\%$ and $98.49\%$, which require model to generalize well to unseen concepts. We use BLEU-3/4~\cite{papineni2002bleu}, METEOR~\cite{banerjee2005meteor}, ROUGE-2/L~\cite{lin2003automatic}, CIDEr~\cite{vedantam2015cider}, and SPICE~\cite{anderson2016spice} as evaluation metrics.

We employ \emph{BART} Large model~\cite{lewis2019bart} as the pretrained generation model. We adopt cross-entropy loss with 0.1 label-smoothing penalty. The $\lambda$ in Equation~\ref{overall_loss} is 1.0. We use inverse-sqrt learning rate scheduler with 500 warmup steps, the learning rate, max-tokens per batch and max updates are 4e-5, 1024 and 5k. The dropout rate is 0.1. We set the standard deviation of initialization in group embedding, scaling module and prototype position indicator to 5e-3. The optimizer of model is Adam~\cite{kingma2014adam} with $\beta_{1}=0.9$ and $\beta_{2}=0.999$. During decoding, the size of beam search is 5.

\subsection{Overall Performance}
To compare our methods with baseline methods, we classify them into three groups. 

\paragraph{Group 1}Models without pretraining. \emph{bRNN-CopyNet} and \emph{Trans-CopyNet} are based on the best popular architecture Bidirectional RNNs and Transformers~\cite{vaswani2017attention} with attention and copy mechanism~\cite{gu2016incorporating}. \emph{MeanPooling-CopyNet} is employed to deal with the influence of the concept ordering in the sequential based methods, where the input concepts is randomly permuted multiple times and decoding is with a mean pooling based MLP network. Levenshtein Transformer~\cite{gu2019levenshtein} is an edit-based non-autoregressive generation model, where the generated sentences go through multiple refinement.
 

\paragraph{Group 2} Pretrained language generation models including GPT-2~\cite{radford2019language}, UniLM~\cite{dong2019unified}, UniLM-v2~\cite{bao2020unilmv2}, BERT-Gen~\cite{bao2020unilmv2}, BART~\cite{lewis2019bart}, and T5~\cite{raffel2019exploring}. All these models are trained with a seq2seq format.


\paragraph{Group 3} Methods proposed in this work which are based on external knowledge dataset $\mathcal{D}_{in}$ and $\mathcal{D}_{out}$. \emph{Retrieve}$_{\mathcal{D}_{*}},*\in\{in,out\}$ take the prototype retrieved from $\mathcal{D}_{in}$ and $\mathcal{D}_{out}$ as the hypothesises. \emph{BART}$_{\mathcal{D}_{*}},*\in\{in,out\}$ feed the concatenation of concepts and prototype retrieved from $\mathcal{D}_{in}$ and $\mathcal{D}_{out}$ into \emph{BART}.  \emph{EKI-BART}$_{\mathcal{D}_{*}},*\in\{in,out\}$ apply our proposed model in $\mathcal{D}_{in}$ and $\mathcal{D}_{out}$, respectively.

\vspace{12pt}
We list the performance of different models in Table~\ref{OverallPerformance}. According to the results, we have several findings.
\begin{enumerate}[-]
\itemsep-0.2em
    \item Performance of pretrained models are far better than these models without pretraining, which demonstrates that training from scratch with data in CommonGen does not suffice for the concepts-based generation. Models pretrained in large scale corpus do learn more knowledge that would contribute to the generation.
    \item The models with prototype retrieved from $\mathcal{D}_{in}$ are better than those with $\mathcal{D}_{out}$, this shows that in-domain dataset $\mathcal{D}_{in}$ consisting of daily scenario descriptions provided more relevant and high-quality prototype than $\mathcal{D}_{out}$.
    \item  \emph{BART}$_{\mathcal{D}_{*}}$ and \emph{EKI-BART}$_{\mathcal{D}_{*}}$, $*\in\{in,out\}$ both outperform the \emph{BART} baseline, which indicates that introduce external text knowledge as prototype would contribute to the concept based generation. Prototype provides effective scenario bias to find out the reasonable concept combination for the generation.  
    \item  \emph{EKI-BART}$_{\mathcal{D}_{in}}$ and \emph{EKI-BART}$_{\mathcal{D}_{out}}$ both perform better than their count-part models \emph{BART}$_{\mathcal{D}_{in}}$ and \emph{BART}$_{\mathcal{D}_{out}}$. Our model is able to achieve improvement in both in-domain and out-of-domain datasets. 
\end{enumerate}


\subsection{Ablation Study}
In this section, we perform ablation study on the development and test dataset to dive into the effectiveness of different components in our model. We use the $\mathcal{D}_{in}$ as knowledge dataset. The baseline is the retrieval-based model and the pretrained based model without any prototype text. We use \emph{GE}, \emph{SM} and \emph{PPI} to denote group embedding, scaling module and prototype position indicator, respectively. Several findings stand out:
\begin{enumerate}[-]
\itemsep-0.2em
\item \emph{BART$_{\mathcal{D}_{in}}$+GE} and \emph{BART$_{\mathcal{D}_{in}}$+GE+SM} outperform \emph{BART$_{\mathcal{D}_{in}}$} and \emph{BART$_{\mathcal{D}_{in}}$+SM}, respectively. This shows that the group embedding which better distinguish concept and prototype is benefit to the generation.
\item \emph{BART$_{\mathcal{D}_{in}}$+SM} and \emph{BART$_{\mathcal{D}_{in}}$+GE+SM} perform better than \emph{BART$_{\mathcal{D}_{in}}$} and \emph{BART$_{\mathcal{D}_{in}}$+GE}, respectively. This verifies the effectiveness of scaling module that better discriminate the noises and effective concepts in retrieved prototypes.
\item \emph{BART$_{\mathcal{D}_{in}}$+GE+SM+PPI} performs better than \emph{BART$_{\mathcal{D}_{in}}$+GE+SM}, achieving 0.7 and 0.8 BLEU-3 improvement on development and test dataset. This demonstrates that informing decoder of the distance from each token to concepts would better identify these important factors in prototype.
\end{enumerate}

\begin{table} [!th]
\begin{center}
\begin{tabular}{l|ccc|ccc}
\midrule[1.0pt]
\multirow{2}{*}{Model}&\multicolumn{3}{c|}{dev} &\multicolumn{3}{c}{test} \\
&\multicolumn{2}{c}{BLEU-3/4} &CIDEr &\multicolumn{2}{c}{BLEU-3/4} &CIDEr \\
\midrule[1.0pt]
\emph{Retrieve} &35.30 &26.70 &13.50 &35.00 &26.40 &12.91 \\
\emph{BART$_{\mathcal{D}_{in}}$} &41.60 &32.20 &16.25 &42.20 &32.40 &16.43 \\
\emph{BART$_{\mathcal{D}_{in}}$+GE} &43.10 &33.40 &16.52 &43.70 &33.90 &16.88 \\
\emph{BART$_{\mathcal{D}_{in}}$+SM} &44.10 &34.20 & 17.06 &44.70 &34.80 &17.11 \\
\emph{BART$_{\mathcal{D}_{in}}$+GE+SM} &44.70 &35.00 &17.20 &45.20 &35.50 &17.40 \\
\midrule[1.0pt]
\emph{BART$_{\mathcal{D}_{in}}$+GE+SM+PPI} &\textbf{45.40} &\textbf{35.60} &\textbf{17.60} &\textbf{46.00} &\textbf{36.10}  &\textbf{17.80} \\
\midrule[1.0pt]
\end{tabular}
\end{center}
\caption{The performance of different modules combination with the external text knowledge dataset $\mathcal{D}_{in}$. } 
\label{AblationStudy}
\end{table}

\subsection{Effectiveness of Scaling Module}
Here, we compare our scaling module with hard mask strategy. We have two implementations of hard masking:
\begin{enumerate}[-]
\itemsep-0.2em
\item \emph{HM}$_{1}$: After encoding, we mask the output states of $\mathcal{O}$ and only keep that of $\mathcal{C}$. 
\item \emph{HM}$_{2}$: We mask these states of tokens $s_{v}\in\mathcal{O},\forall c\in\mathcal{C},c\neq s_{v}$.
\item \emph{SM}$_{0}$: We remove the encoder classification mechanism from scaling module.
\end{enumerate}
 The experiment is conducted in $\mathcal{D}_{in}$ and we list the results in Table~\ref{hardmaskresult}.

From the results in Table~\ref{hardmaskresult}, first, we can see that 
the last four models perform better than the counter-part model \emph{BART$_{\mathcal{D}_{in}}$+GE}, this verifies that it is beneficial to remove the noises in prototype. Second,
performance of \emph{BART$_{\mathcal{D}_{in}}$+GE+SM} and \emph{BART$_{\mathcal{D}_{in}}$+GE+SM$_{0}$} are better than both \emph{BART$_{\mathcal{D}_{in}}$+GE+HM$_{1}$} and \emph{BART$_{\mathcal{D}_{in}}$+GE+HM$_{2}$}, this indicates that our scaling module is better than the hard masking strategies \emph{HM$_{1}$} and \emph{HM$_{2}$}. This phenomenon demonstrates that there exists more effective additional concepts besides those concept tokens in prototype that would contribute to build the target scene, directly masking these tokens would block the generator receiving these additional information, but our scaling module is able to keep these additional information. Third, \emph{BART$_{\mathcal{D}_{in}}$+GE+SM} outperforms \emph{BART$_{\mathcal{D}_{in}}$+GE+SM$_{0}$}, this shows that injecting the prior knowledge of prototype in scaling module would boost the performance of scaling module.

\begin{table} [!th]
\begin{center}
\begin{tabular}{l|ccc|ccc}
\midrule[1.0pt]
\multirow{2}{*}{Model}&\multicolumn{3}{c|}{dev} &\multicolumn{3}{c}{test} \\
&\multicolumn{2}{c}{BLEU-3/4} &CIDEr &\multicolumn{2}{c}{BLEU-3/4} &CIDEr \\
\midrule[1.0pt]
\emph{BART$_{\mathcal{D}_{in}}$+GE} &43.10 &33.40 &16.52 &43.70 &33.90 &16.88 \\
\emph{BART$_{\mathcal{D}_{in}}$+GE+HM$_{1}$} &43.90 &34.00 &16.84 &44.60 &34.50 &16.96\\
\emph{BART$_{\mathcal{D}_{in}}$+GE+HM$_{2}$} &44.00 &34.10 & 17.01 &44.90 &34.50 &17.21 \\
\emph{BART$_{\mathcal{D}_{in}}$+GE+SM$_{0}$} &44.10 &34.20 &17.06 &44.70 &34.80 &17.31 \\
\emph{BART$_{\mathcal{D}_{in}}$+GE+SM} &\textbf{44.70} &\textbf{35.00} &\textbf{17.22} &\textbf{45.40} &\textbf{35.60} &\textbf{17.52} \\
\midrule[1.0pt]
\end{tabular}
\end{center}
\caption{The performance on plain text knowledge dataset $\mathcal{D}_{in}$. \emph{GE}, \emph{SM} and \emph{PPI} are short for group embedding, scaling module and prototype position indicator, respectively.} 
\label{hardmaskresult}
\end{table}

\subsection{Missing Concept Number in Generation}
\begin{figure}[!th]
    \centering
    \includegraphics[width = 3.5in]{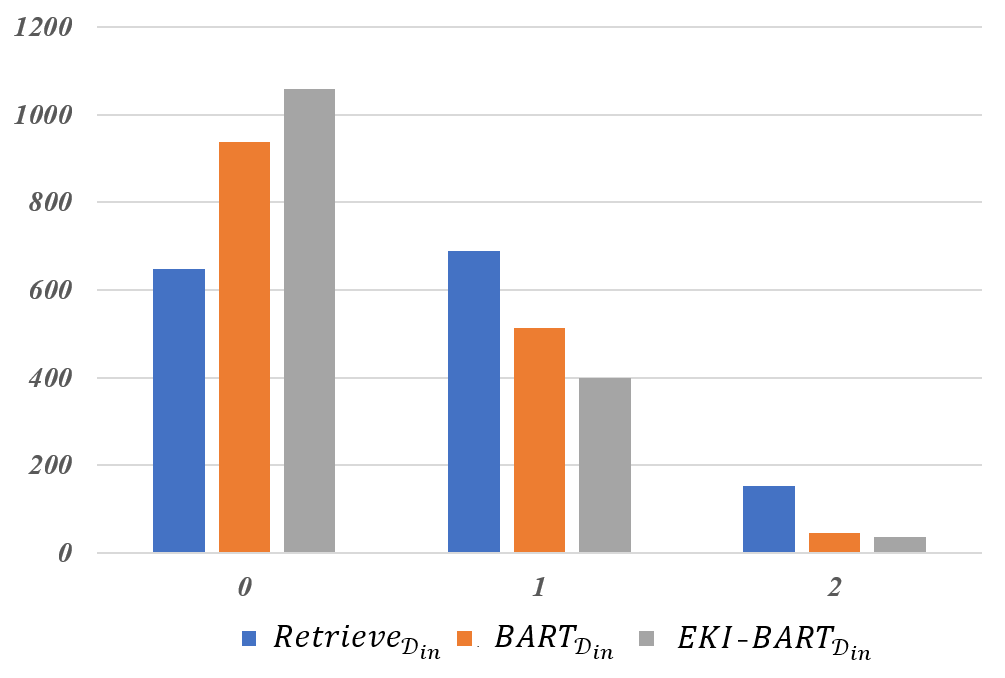}
	\caption{Number of missing concepts in \emph{Retrieve}$_{\mathcal{D}_{in}}$, \emph{BART}$_{\mathcal{D}_{in}}$ and \emph{EKI-BART}$_{\mathcal{D}_{in}}$. X-axis is the missing concept number in each sentence, Y-axis is the instance number in the test set of CommonGen.}
	\label{MissingConceptNum}
\end{figure}

CommonGen aims to generate scenario description that contains all of these provided concepts. If the model is able to find out the most plausible scene with these concepts, these would be no concepts missing in the generated sentence. To check whether our model is able to find out better scene on the basis of retrieved prototype, thus we compare the number of missing concepts in \emph{Retrieve}$_{\mathcal{D}_{in}}$, \emph{BART}$_{\mathcal{D}_{in}}$ and \emph{EKI-BART}$_{\mathcal{D}_{in}}$ and list the results in Figure~\ref{MissingConceptNum}. 

From Figure~\ref{MissingConceptNum}, there are another 300+ instances with no concepts missing in generation of \emph{BART}$_{\mathcal{D}_{in}}$ and \emph{EKI-BART}$_{\mathcal{D}_{in}}$ compared to \emph{Retrieve}$_{\mathcal{D}_{in}}$, we easily conclude that the two models are able to inject more concepts into the retrieved prototype and further utilize the prototype knowledge to generate a more plausible sentence. We also notice that the number of instances with no concept missing of \emph{EKI-BART}$_{\mathcal{D}_{in}}$ is more than that of \emph{BART}$_{\mathcal{D}_{in}}$, which shows that \emph{BART}$_{\mathcal{D}_{in}}$ is more likely to ignore the provided concepts than \emph{EKI-BART}$_{\mathcal{D}_{in}}$ and being dominated by noises in prototype. This verifies that the ability of \emph{EKI-BART}$_{\mathcal{D}_{in}}$ in dealing with prototype noises is stronger than \emph{BART}$_{\mathcal{D}_{in}}$, and removing these noises is useful to construct a more plausible scenario. 

\section{Related Work}
\subsection{Commonsense Reasoning} 
Recently, there are emerging works to investigate machine commonsense reasoning ability. ATOMIC~\cite{sap2019atomic}, Event2Mind~\cite{rashkin2018event2mind}, MCScript 2.0~\cite{ostermann2019mcscript2}, SWAG~\cite{zellers2018swag}, HellaSWAG~\cite{zellers2019hellaswag}, Story Cloze Test~\cite{mostafazadeh2017lsdsem}, CommonsenseQA~\cite{talmor2018commonsenseqa} and CommonGen~\cite{lin2019comgen} are released to 
reason over external knowledge besides the inputs for question answering or generation. Rajani et al.~\shortcite{rajani2019explain} explore adding human-written explanations to solve the problem. Lin et al.~\shortcite{lin-etal-2019-kagnet} construct schema graphs from ConceptNet to reason over relevant commonsense knowledge. lv et al.~\shortcite{lv2020graph} focus on automatically extracting evidence from heterogeneous external knowledge and reasoning over the extracted evidence to study this problem. Considering quite a few relationships about these concepts require a variety of background knowledge such as spatial relations, object properties, physical rules, temporal event knowledge, social conventions, etc., which may not be recorded in any existing knowledge bases, this paper focuses on retrieving knowledge from plain text in order to introduce scenario bias for concepts-set based generation. 

\subsection{retrieve-and-generation} 
The retrieve-and-generation approaches are developed for many tasks, including dialogue generation~\cite{weston-etal-2018-retrieve,song2016two}, language modeling~\cite{guu-etal-2018-generating}, code generation~\cite{hashimoto2018retrieve} and text summarization~\cite{rush-etal-2015-neural,cao-etal-2018-retrieve,peng2019text}. Ji et al.~\shortcite{ji2014information} and Yan et al.~\shortcite{yan2016learning} focuses on prototype ranking in the retrieval-based model but they do not edit these retrieved prototypes. Re3Sum~\cite{cao2018retrieve} is an LSTM-based model developed under the retrieve-and-generation framework that retrieves multiple headlines and pick the single best retrieved headline, then edit. Hashimoto et al.~\cite{hashimoto2018retrieve} Hossain et al.~\shortcite{hossain2020simple} presents a framework that retrieve, generation and rerank on the basis of BERT~\cite{devlin2018bert}, but they do not deal with prototype noise in an explicit manner. Song et al.~\shortcite{song2016two} introduces an extra encoder for the retrieved response, and the output of the encoder, together with that of the query encoder, is utilized to feed the decoder. Weston et al.~\shortcite{weston-etal-2018-retrieve} simply concatenates the original query and the retrieved response as the input to the encoder. Instead of solely using the retrieved response, Wu et al.~\shortcite{wu2019response} further introduces to encode the lexical differences between the current query and the retrieved query. Pandey et al.~\shortcite{pandey2018exemplar} proposes to weight different training instances by context similarity. Different from these works, we explore the retrieve-and-generation framework on the basis of pretrained encoder-decoder model, and identify the importance of each token in prototype in a more fine-grained manner.

\section{Conclusion and Future Work}
\label{SectionConclusion}
In this paper, we proposed an enhanced retrieve-and-generation model for commonsense generation. The key of CommonGen is to identify the priority of the scene based on the concept combination. We have scaling module to softly reduce the impact of prototype noises on generation and prototype position indicator to help decoder to learn the prototype scenario better. Both our retrieve-and-generation model with in-domain and out-of-domain datasets achieve better performance than retrieval and pretrained encoder-decoder methods. In future, we plan to build the relationship of these concepts in a more structure manner.

\section{Acknowledgments}
This work is partically supported by National Natural Science Foundation of China (No. 71991471), Science and Technology Commission of Shanghai Municipality Grant (No.20dz1200600, No.18DZ1201000, 17JC1420200).

\bibliographystyle{coling}
\bibliography{coling2020}

\begin{thebibliography}{}

\bibitem[\protect\citename{Anderson \bgroup et al.\egroup
  }2016]{anderson2016spice}
Peter Anderson, Basura Fernando, Mark Johnson, and Stephen Gould.
\newblock 2016.
\newblock Spice: Semantic propositional image caption evaluation.
\newblock In {\em European Conference on Computer Vision}, pages 382--398.
  Springer.

\bibitem[\protect\citename{Banerjee and Lavie}2005]{banerjee2005meteor}
Satanjeev Banerjee and Alon Lavie.
\newblock 2005.
\newblock {METEOR}: An automatic metric for {MT} evaluation with improved
  correlation with human judgments.
\newblock In {\em Proceedings of the ACL Workshop on Intrinsic and Extrinsic
  Evaluation Measures for Machine Translation and/or Summarization}, pages
  65--72.

\bibitem[\protect\citename{Bao \bgroup et al.\egroup }2020]{bao2020unilmv2}
Hangbo Bao, Li~Dong, Furu Wei, Wenhui Wang, Nan Yang, Xiaodong Liu, Yu~Wang,
  Songhao Piao, Jianfeng Gao, Ming Zhou, et~al.
\newblock 2020.
\newblock Unilmv2: Pseudo-masked language models for unified language model
  pre-training.
\newblock {\em arXiv preprint arXiv:2002.12804}.

\bibitem[\protect\citename{Bowman \bgroup et al.\egroup }2015]{snliemnlp2015}
Samuel~R. Bowman, Gabor Angeli, Christopher Potts, and Christopher~D. Manning.
\newblock 2015.
\newblock A large annotated corpus for learning natural language inference.
\newblock In {\em Proceedings of the 2015 Conference on Empirical Methods in
  Natural Language Processing (EMNLP)}.

\bibitem[\protect\citename{Cao \bgroup et al.\egroup
  }2018a]{cao-etal-2018-retrieve}
Ziqiang Cao, Wenjie Li, Sujian Li, and Furu Wei.
\newblock 2018a.
\newblock Retrieve, rerank and rewrite: Soft template based neural
  summarization.
\newblock In {\em Proceedings of the 56th Annual Meeting of the Association for
  Computational Linguistics (Volume 1: Long Papers)}, pages 152--161,
  Melbourne, Australia, July. Association for Computational Linguistics.

\bibitem[\protect\citename{Cao \bgroup et al.\egroup }2018b]{cao2018retrieve}
Ziqiang Cao, Wenjie Li, Sujian Li, and Furu Wei.
\newblock 2018b.
\newblock Retrieve, rerank and rewrite: Soft template based neural
  summarization.
\newblock In {\em Proceedings of the 56th Annual Meeting of the Association for
  Computational Linguistics (Volume 1: Long Papers)}, pages 152--161.

\bibitem[\protect\citename{Devlin \bgroup et al.\egroup }2018]{devlin2018bert}
Jacob Devlin, Ming-Wei Chang, Kenton Lee, and Kristina Toutanova.
\newblock 2018.
\newblock Bert: Pre-training of deep bidirectional transformers for language
  understanding.
\newblock {\em arXiv preprint arXiv:1810.04805}.

\bibitem[\protect\citename{Dong \bgroup et al.\egroup }2019]{dong2019unified}
Li~Dong, Nan Yang, Wenhui Wang, Furu Wei, Xiaodong Liu, Yu~Wang, Jianfeng Gao,
  Ming Zhou, and Hsiao-Wuen Hon.
\newblock 2019.
\newblock Unified language model pre-training for natural language
  understanding and generation.
\newblock In {\em Advances in Neural Information Processing Systems}, pages
  13042--13054.

\bibitem[\protect\citename{Gu \bgroup et al.\egroup }2016]{gu2016incorporating}
Jiatao Gu, Zhengdong Lu, Hang Li, and Victor~OK Li.
\newblock 2016.
\newblock Incorporating copying mechanism in sequence-to-sequence learning.
\newblock {\em arXiv preprint arXiv:1603.06393}.

\bibitem[\protect\citename{Gu \bgroup et al.\egroup }2019]{gu2019levenshtein}
Jiatao Gu, Changhan Wang, and Junbo Zhao.
\newblock 2019.
\newblock Levenshtein transformer.
\newblock In {\em Advances in Neural Information Processing Systems}, pages
  11179--11189.

\bibitem[\protect\citename{Guu \bgroup et al.\egroup
  }2018]{guu-etal-2018-generating}
Kelvin Guu, Tatsunori~B. Hashimoto, Yonatan Oren, and Percy Liang.
\newblock 2018.
\newblock Generating sentences by editing prototypes.
\newblock {\em Transactions of the Association for Computational Linguistics},
  6:437--450.

\bibitem[\protect\citename{Hashimoto \bgroup et al.\egroup
  }2018]{hashimoto2018retrieve}
Tatsunori~B Hashimoto, Kelvin Guu, Yonatan Oren, and Percy~S Liang.
\newblock 2018.
\newblock A retrieve-and-edit framework for predicting structured outputs.
\newblock In {\em Advances in Neural Information Processing Systems}, pages
  10052--10062.

\bibitem[\protect\citename{Hossain \bgroup et al.\egroup
  }2020]{hossain2020simple}
Nabil Hossain, Marjan Ghazvininejad, and Luke Zettlemoyer.
\newblock 2020.
\newblock Simple and effective retrieve-edit-rerank text generation.
\newblock In {\em Proceedings of the 58th Annual Meeting of the Association for
  Computational Linguistics}, pages 2532--2538.

\bibitem[\protect\citename{Ji \bgroup et al.\egroup }2014]{ji2014information}
Zongcheng Ji, Zhengdong Lu, and Hang Li.
\newblock 2014.
\newblock An information retrieval approach to short text conversation.
\newblock {\em arXiv preprint arXiv:1408.6988}.

\bibitem[\protect\citename{Kingma and Ba}2014]{kingma2014adam}
Diederik~P Kingma and Jimmy Ba.
\newblock 2014.
\newblock Adam: A method for stochastic optimization.
\newblock {\em arXiv preprint arXiv:1412.6980}.

\bibitem[\protect\citename{Krishna \bgroup et al.\egroup
  }2017]{krishna2017dense}
Ranjay Krishna, Kenji Hata, Frederic Ren, Li~Fei-Fei, and Juan Carlos~Niebles.
\newblock 2017.
\newblock Dense-captioning events in videos.
\newblock In {\em Proceedings of the IEEE international conference on computer
  vision}, pages 706--715.

\bibitem[\protect\citename{Lewis \bgroup et al.\egroup }2019]{lewis2019bart}
Mike Lewis, Yinhan Liu, Naman Goyal, Marjan Ghazvininejad, Abdelrahman Mohamed,
  Omer Levy, Ves Stoyanov, and Luke Zettlemoyer.
\newblock 2019.
\newblock Bart: Denoising sequence-to-sequence pre-training for natural
  language generation, translation, and comprehension.
\newblock {\em arXiv preprint arXiv:1910.13461}.

\bibitem[\protect\citename{Lin and Hovy}2003]{lin2003automatic}
Chin-Yew Lin and Eduard Hovy.
\newblock 2003.
\newblock Automatic evaluation of summaries using {N}-gram co-occurrence
  statistics.
\newblock In {\em Proceedings of the 2003 Conference of the North American
  Chapter of the Association for Computational Linguistics on Human Language
  Technology-Volume 1}, pages 71--78. Association for Computational
  Linguistics.

\bibitem[\protect\citename{Lin \bgroup et al.\egroup
  }2019a]{lin-etal-2019-kagnet}
Bill~Yuchen Lin, Xinyue Chen, Jamin Chen, and Xiang Ren.
\newblock 2019a.
\newblock {K}ag{N}et: Knowledge-aware graph networks for commonsense reasoning.
\newblock In {\em Proceedings of the 2019 Conference on Empirical Methods in
  Natural Language Processing and the 9th International Joint Conference on
  Natural Language Processing (EMNLP-IJCNLP)}, pages 2829--2839, Hong Kong,
  China, November. Association for Computational Linguistics.

\bibitem[\protect\citename{Lin \bgroup et al.\egroup }2019b]{lin2019comgen}
Bill~Yuchen Lin, Ming Shen, Wangchunshu Zhou, Pei Zhou, Chandra Bhagavatula,
  Yejin Choi, and Xiang Ren.
\newblock 2019b.
\newblock Commongen: A constrained text generation challenge for generative
  commonsense reasoning.
\newblock {\em CoRR}, abs/1911.03705.

\bibitem[\protect\citename{Lv \bgroup et al.\egroup }2020]{lv2020graph}
Shangwen Lv, Daya Guo, Jingjing Xu, Duyu Tang, Nan Duan, Ming Gong, Linjun
  Shou, Daxin Jiang, Guihong Cao, and Songlin Hu.
\newblock 2020.
\newblock Graph-based reasoning over heterogeneous external knowledge for
  commonsense question answering.
\newblock In {\em AAAI}, pages 8449--8456.

\bibitem[\protect\citename{Mostafazadeh \bgroup et al.\egroup
  }2017]{mostafazadeh2017lsdsem}
Nasrin Mostafazadeh, Michael Roth, Annie Louis, Nathanael Chambers, and James
  Allen.
\newblock 2017.
\newblock Lsdsem 2017 shared task: The story cloze test.
\newblock In {\em Proceedings of the 2nd Workshop on Linking Models of Lexical,
  Sentential and Discourse-level Semantics}, pages 46--51.

\bibitem[\protect\citename{Ostermann \bgroup et al.\egroup
  }2019]{ostermann2019mcscript2}
Simon Ostermann, Michael Roth, and Manfred Pinkal.
\newblock 2019.
\newblock Mcscript2. 0: A machine comprehension corpus focused on script events
  and participants.
\newblock {\em arXiv preprint arXiv:1905.09531}.

\bibitem[\protect\citename{Pandey \bgroup et al.\egroup
  }2018]{pandey2018exemplar}
Gaurav Pandey, Danish Contractor, Vineet Kumar, and Sachindra Joshi.
\newblock 2018.
\newblock Exemplar encoder-decoder for neural conversation generation.
\newblock In {\em Proceedings of the 56th Annual Meeting of the Association for
  Computational Linguistics (Volume 1: Long Papers)}, pages 1329--1338.

\bibitem[\protect\citename{Papineni \bgroup et al.\egroup
  }2002]{papineni2002bleu}
Kishore Papineni, Salim Roukos, Todd Ward, and Wei-Jing Zhu.
\newblock 2002.
\newblock {BLEU}: a method for automatic evaluation of machine translation.
\newblock In {\em Proceedings of the 40th Annual Meeting on Association for
  Computational Linguistics}, pages 311--318. Association for Computational
  Linguistics.

\bibitem[\protect\citename{Peng \bgroup et al.\egroup }2019]{peng2019text}
Hao Peng, Ankur~P Parikh, Manaal Faruqui, Bhuwan Dhingra, and Dipanjan Das.
\newblock 2019.
\newblock Text generation with exemplar-based adaptive decoding.
\newblock {\em arXiv preprint arXiv:1904.04428}.

\bibitem[\protect\citename{Radford \bgroup et al.\egroup
  }2019]{radford2019language}
Alec Radford, Jeffrey Wu, Rewon Child, David Luan, Dario Amodei, and Ilya
  Sutskever.
\newblock 2019.
\newblock Language models are unsupervised multitask learners.
\newblock {\em OpenAI Blog}, 1(8):9.

\bibitem[\protect\citename{Raffel \bgroup et al.\egroup
  }2019]{raffel2019exploring}
Colin Raffel, Noam Shazeer, Adam Roberts, Katherine Lee, Sharan Narang, Michael
  Matena, Yanqi Zhou, Wei Li, and Peter~J Liu.
\newblock 2019.
\newblock Exploring the limits of transfer learning with a unified text-to-text
  transformer.
\newblock {\em arXiv preprint arXiv:1910.10683}.

\bibitem[\protect\citename{Rajani \bgroup et al.\egroup
  }2019]{rajani2019explain}
Nazneen~Fatema Rajani, Bryan McCann, Caiming Xiong, and Richard Socher.
\newblock 2019.
\newblock Explain yourself! leveraging language models for commonsense
  reasoning.
\newblock {\em arXiv preprint arXiv:1906.02361}.

\bibitem[\protect\citename{Rashkin \bgroup et al.\egroup
  }2018]{rashkin2018event2mind}
Hannah Rashkin, Maarten Sap, Emily Allaway, Noah~A Smith, and Yejin Choi.
\newblock 2018.
\newblock Event2mind: Commonsense inference on events, intents, and reactions.
\newblock {\em arXiv preprint arXiv:1805.06939}.

\bibitem[\protect\citename{Rush \bgroup et al.\egroup
  }2015]{rush-etal-2015-neural}
Alexander~M. Rush, Sumit Chopra, and Jason Weston.
\newblock 2015.
\newblock A neural attention model for abstractive sentence summarization.
\newblock In {\em Proceedings of the 2015 Conference on Empirical Methods in
  Natural Language Processing}, pages 379--389, Lisbon, Portugal, September.
  Association for Computational Linguistics.

\bibitem[\protect\citename{Sap \bgroup et al.\egroup }2019]{sap2019atomic}
Maarten Sap, Ronan Le~Bras, Emily Allaway, Chandra Bhagavatula, Nicholas
  Lourie, Hannah Rashkin, Brendan Roof, Noah~A Smith, and Yejin Choi.
\newblock 2019.
\newblock Atomic: An atlas of machine commonsense for if-then reasoning.
\newblock In {\em Proceedings of the AAAI Conference on Artificial
  Intelligence}, volume~33, pages 3027--3035.

\bibitem[\protect\citename{Song \bgroup et al.\egroup }2016]{song2016two}
Yiping Song, Rui Yan, Xiang Li, Dongyan Zhao, and Ming Zhang.
\newblock 2016.
\newblock Two are better than one: An ensemble of retrieval-and
  generation-based dialog systems.
\newblock {\em arXiv preprint arXiv:1610.07149}.

\bibitem[\protect\citename{Talmor \bgroup et al.\egroup
  }2018]{talmor2018commonsenseqa}
Alon Talmor, Jonathan Herzig, Nicholas Lourie, and Jonathan Berant.
\newblock 2018.
\newblock Commonsenseqa: A question answering challenge targeting commonsense
  knowledge.
\newblock {\em arXiv preprint arXiv:1811.00937}.

\bibitem[\protect\citename{Vaswani \bgroup et al.\egroup
  }2017]{vaswani2017attention}
Ashish Vaswani, Noam Shazeer, Niki Parmar, Jakob Uszkoreit, Llion Jones,
  Aidan~N Gomez, \L~ukasz Kaiser, and Illia Polosukhin.
\newblock 2017.
\newblock Attention is all you need.
\newblock In I.~Guyon, U.~V. Luxburg, S.~Bengio, H.~Wallach, R.~Fergus,
  S.~Vishwanathan, and R.~Garnett, editors, {\em Advances in Neural Information
  Processing Systems 30}, pages 5998--6008. Curran Associates, Inc.

\bibitem[\protect\citename{Vedantam \bgroup et al.\egroup
  }2015]{vedantam2015cider}
Ramakrishna Vedantam, C~Lawrence~Zitnick, and Devi Parikh.
\newblock 2015.
\newblock Cider: Consensus-based image description evaluation.
\newblock In {\em Proceedings of the IEEE Conference on Computer Vision and
  Pattern Recognition}, pages 4566--4575.

\bibitem[\protect\citename{Wang \bgroup et al.\egroup }2019]{wang2019vatex}
Xin Wang, Jiawei Wu, Junkun Chen, Lei Li, Yuan-Fang Wang, and William~Yang
  Wang.
\newblock 2019.
\newblock Vatex: A large-scale, high-quality multilingual dataset for
  video-and-language research.
\newblock In {\em Proceedings of the IEEE International Conference on Computer
  Vision}, pages 4581--4591.

\bibitem[\protect\citename{Weston \bgroup et al.\egroup
  }2018]{weston-etal-2018-retrieve}
Jason Weston, Emily Dinan, and Alexander Miller.
\newblock 2018.
\newblock Retrieve and refine: Improved sequence generation models for
  dialogue.
\newblock In {\em Proceedings of the 2018 {EMNLP} Workshop {SCAI}: The 2nd
  International Workshop on Search-Oriented Conversational {AI}}, pages 87--92,
  Brussels, Belgium, October. Association for Computational Linguistics.

\bibitem[\protect\citename{Wu \bgroup et al.\egroup }2019]{wu2019response}
Yu~Wu, Furu Wei, Shaohan Huang, Yunli Wang, Zhoujun Li, and Ming Zhou.
\newblock 2019.
\newblock Response generation by context-aware prototype editing.
\newblock In {\em Proceedings of the AAAI Conference on Artificial
  Intelligence}, volume~33, pages 7281--7288.

\bibitem[\protect\citename{Yan \bgroup et al.\egroup }2016]{yan2016learning}
Rui Yan, Yiping Song, and Hua Wu.
\newblock 2016.
\newblock Learning to respond with deep neural networks for retrieval-based
  human-computer conversation system.
\newblock In {\em Proceedings of the 39th International ACM SIGIR conference on
  Research and Development in Information Retrieval}, pages 55--64.

\bibitem[\protect\citename{Zellers \bgroup et al.\egroup
  }2018]{zellers2018swag}
Rowan Zellers, Yonatan Bisk, Roy Schwartz, and Yejin Choi.
\newblock 2018.
\newblock Swag: A large-scale adversarial dataset for grounded commonsense
  inference.
\newblock {\em arXiv preprint arXiv:1808.05326}.

\bibitem[\protect\citename{Zellers \bgroup et al.\egroup
  }2019]{zellers2019hellaswag}
Rowan Zellers, Ari Holtzman, Yonatan Bisk, Ali Farhadi, and Yejin Choi.
\newblock 2019.
\newblock Hellaswag: Can a machine really finish your sentence?
\newblock {\em arXiv preprint arXiv:1905.07830}.

\end{thebibliography}

\end{document}